\documentclass[11pt]{article}

\usepackage[margin=1in]{geometry}
\usepackage{amsmath, amssymb, amsthm}
\usepackage{graphicx}
\usepackage{xcolor}
\usepackage{hyperref}
\hypersetup{
  colorlinks=true,
  linkcolor=black,
  citecolor=black,
  urlcolor=blue
}
\usepackage{newtxtext}
\usepackage{newtxmath}
\usepackage{enumitem}
\setlist[itemize]{nosep,leftmargin=*}
\usepackage{parskip}
\usepackage{tikz}
\usetikzlibrary{arrows.meta, positioning, shapes.multipart, fit, calc, backgrounds}
\usepackage{float}
\title{SciNets: Graph-Constrained Multi-Hop Reasoning for Scientific Literature Synthesis}
\author{Sauhard Dubey \\
Jaypee Institute of Information Technology, Noida \\
\texttt{sauhard.research@gmail.com}}
\date{} 

\makeatletter
\g@addto@macro\normalsize{%
  \setlength\abovedisplayskip{5pt}
  \setlength\belowdisplayskip{5pt}
  \setlength\abovedisplayshortskip{3pt}
  \setlength\belowdisplayshortskip{3pt}
}
\makeatother

\begin{document}

\maketitle

\begin{abstract}
\noindent
Cross-domain scientific synthesis requires connecting mechanistic explanations across fragmented literature, a capability that remains challenging for both retrieval-based systems and unconstrained language models. While recent work has applied large language models to scientific summarization and question answering, these approaches provide limited control over reasoning depth and structural grounding. We frame mechanistic synthesis as a graph-constrained multi-hop reasoning problem over literature-derived concept graphs. Given a scientific query and a compact, query-local corpus, SciNets constructs a directed concept graph and synthesizes mechanistic explanations by identifying multi-hop reasoning paths that connect concepts that rarely co-occur within individual papers. We systematically compare shortest-path reasoning, $k$-shortest paths with diversity constraints, stochastic random walks, and a retrieval-augmented language model baseline. Rather than evaluating correctness, which is often indeterminate when synthesizing connections across distributed sources, we introduce a behavioral framework that measures symbolic reasoning depth, mechanistic diversity, and grounding stability. Across machine learning, biology, and climate science tasks, explicit graph constraints enable controllable multi-hop reasoning while revealing a consistent trade-off: deeper and more diverse symbolic reasoning increases grounding instability, whereas shortest-path reasoning remains highly stable but structurally conservative. These findings provide a systematic behavioral characterization of the limits and capabilities of current graph–LLM integration for scientific synthesis.
\end{abstract}

\section{Introduction}
Scientific progress increasingly depends on the ability to synthesize mechanistic explanations that connect concepts distributed across a fragmented and heterogeneous literature. Although large language models (LLMs) show strong summarization ability, they often struggle to construct the structured multi-hop reasoning chains required for structured exploratory reasoning and can collapse into shallow statements or ungrounded hallucinations in complex reasoning settings. This study introduces SciNets, a structured literature synthesis system that formulates cross-domain mechanistic synthesis as a Graph-Constrained Reasoning problem over query-local concept graphs derived from the literature. Rather than proposing a traditional accuracy benchmark or evaluating correctness, which is often unknowable when synthesizing connections across distributed sources, we introduce a Behavioral evaluation framework that measures the depth of symbolic reasoning, mechanistic diversity, and grounding stability. Our results show a consistent trade-off: strategies that promote deeper symbolic reasoning increase structural stress and grounding instability, whereas shortest-path reasoning preserves reliability at the cost of exploration breadth.
We formalize this phenomenon using Behavioral metrics and trace evidence, showing that Graph-Constrained Reasoning provides controllable multi-hop exploration while revealing practical limits on grounding stability. This positions SciNets as a structured and measurable approach to AI-assisted scientific exploration, rather than an unstructured language generation system.
\section{Contributions}

This study investigates cross-domain mechanistic synthesis as a structured, Graph-Constrained Reasoning task rather than an unstructured language generation problem. The key contributions of this study are as follows:

\begin{itemize}
    \item \textbf{Problem Formulation.} We formulate cross-domain mechanistic synthesis as a graph-constrained multi-hop reasoning problem over query-local concept graphs, explicitly distinguishing structured synthesis from extractive summarization, question answering, and retrieval-based knowledge aggregation (RKA).

    \item \textbf{Methodological Design.} We propose a structured framework that constructs literature-derived concept graphs and synthesizes mechanistic explanations by identifying Graph-Constrained Reasoning paths using multiple traversal strategies, including shortest-path reasoning, $k$-shortest paths with diversity enforcement, stochastic random walks, and a retrieval-augmented language model baseline.

    \item \textbf{Behavior-Centric Evaluation.} Rather than evaluating scientific correctness, we introduced a Behavioral evaluation framework that characterizes reasoning depth, diversity, stability of grounding, and realization fidelity, enabling a systematic comparison of reasoning strategies.

    \item \textbf{Empirical Insights.} Through cross-domain evaluation, we reveal a consistent trade-off between the richness of symbolic reasoning and grounding stability. Diversity-enhanced strategies surface deeper and more varied reasoning chains but frequently collapse during linguistic realization, whereas shortest-path reasoning provides compact and reliably grounded explanations for model outputs.

\end{itemize}

\section{Problem Formulation}

Cross-domain mechanistic synthesis is treated as a structured reasoning problem over literature-derived concept graphs. Unlike summarization or question answering, the objective is not to restate existing knowledge but to surface mechanistic connections that exist implicitly across fragmented literature through explicit multi-hop reasoning paths.

\subsection{Inputs and Outputs}

\textbf{Input.} The system receives:
\begin{enumerate}
    \item A scientific query $q$ expressed in natural language.
    \item A retrieved corpus of documents $D = \{d_1, d_2, \dots, d_N\}$,
    where each document $d_i$ may contain concepts and relations relevant to $q$.
\end{enumerate}

\textbf{Output.} The output is a ranked set of mechanistic explanations
\[
\mathcal{H} = \{h_1, h_2, \dots, h_K\}
\]
where each explanation $h_k$ is expressed in natural language and corresponds to a multi-hop reasoning chain connecting concepts distributed across the retrieved corpus but rarely co-articulated in the individual papers.

\subsection{Concept Graph Construction}

From the retrieved corpus, we construct a query-local directed concept graph
\[
G = (V, E)
\]
where
\begin{itemize}
    \item $V$ denotes scientific concepts extracted from text
    \item $E \subseteq V \times V$ denotes directed relations inferred from contextual evidence
\end{itemize}

The resulting graph is structured and noisy. This reflects relations present in the retrieved literature instead of an authoritative or curated ontology.

\subsection{Mechanistic Explanations as Multi-Hop Reasoning Paths}

A mechanistic explanation is defined as a reasoning path in the graph
\[
p = (v_0 \rightarrow v_1 \rightarrow \dots \rightarrow v_L)
\]
where
\begin{itemize}
    \item $v_0$ and $v_L$ correspond to query relevant concepts
    \item $L \ge 2$ indicates multi-hop reasoning
    \item intermediate nodes represent documented mediating mechanisms
\end{itemize}

Each path p provides a structured synthesis of how concepts documented in separate papers may be connected mechanistically. Therefore, the synthesis task reduces to identifying informative paths in G that bridge structural gaps in the literature and translating them into coherent natural language explanations.

\subsection{Graph-Constrained Mechanistic Synthesis}

Given a concept graph $G$, the synthesis task is to identify paths that
\begin{itemize}
    \item exhibit sufficient reasoning depth
    \item maintain diversity rather than repeating trivial variations
    \item remain grounded in the graph structure instead of relying on language model inference
\end{itemize}

Different reasoning strategies impose different structural constraints on path selection, resulting in trade-offs among reliability, diversity, and synthesis difficulty.
\section{System Intent and Design Philosophy}

SciNets is explicitly designed as a structured literature synthesis engine rather than a question-answering or retrieval system. The goal is not merely to restate existing knowledge or extract consensus judgments but to surface structurally grounded, multi-hop mechanistic connections that exist implicitly in fragmented literature but are difficult to identify through conventional reading.

Three design principles guided the system:

\textbf{(1) Synthesis over Extraction.}
Scientific insights often require connecting concepts across papers that do not cite each other. SciNets is designed to systematically surface multi-hop mechanistic pathways, examine under-discussed conceptual connections, and synthesize explanations that span structural gaps while remaining anchored to literature-derived structures.

\textbf{(2) Structure over Fluency.}
Instead of relying solely on unstructured generative reasoning, SciNets constrains hypothesis synthesis by using literature-derived concept graphs. This separation between symbolic reasoning and natural language realization enables explicit control over the reasoning depth, interpretability, and analytical evaluation.

\textbf{(3) Exploration over Stability.}
By design, SciNets prioritizes structural diversity and community-bridging reasoning. This intentionally pushes the system into regions of conceptual space that are structurally interesting, but linguistically fragile. As shown in our evaluation, this emphasis leads to measurable tension between symbolic richness and grounded stability.
\section{Methodology}

This section describes the proposed framework for the graph-constrained mechanistic synthesis. The method produces mechanistic explanations by combining structured reasoning over literature-derived concept graphs with a controlled natural language realization. This approach is designed to separate symbolic reasoning from linguistic expression, enabling explicit control over the reasoning structure while allowing independent analysis of Behavioral properties.

\subsection{Pipeline Overview}
\begin{figure}[htbp]
\centering
\begin{tikzpicture}[
node distance=0.85cm,
every node/.style={draw, rounded corners, align=center, minimum width=8cm, minimum height=1.2cm},
arrow/.style={-Latex, thick}
]

\node (query) { \textbf{Scientific Query} \\ \small Natural language research question };
\node (retrieval) [below=of query] { \textbf{Literature Retrieval} \\ \small Query-relevant corpus };
\node (graph) [below=of retrieval] { \textbf{Concept Graph Construction} \\ \small Directed query-local concept graph };
\node (reason) [below=of graph] { \textbf{Graph-Constrained Reasoning} \\ \small Shortest-path, $k$-shortest paths, random walks };
\node (synthesis) [below=of reason] { \textbf{Explanation Synthesis} \\ \small Multi-hop mechanistic explanations };

\draw[arrow] (query) -- (retrieval);
\draw[arrow] (retrieval) -- (graph);
\draw[arrow] (graph) -- (reason);
\draw[arrow] (reason) -- (synthesis);

\end{tikzpicture}
\caption{Overview of the proposed graph-constrained mechanistic synthesis pipeline}
\label{fig:pipeline}
\end{figure}
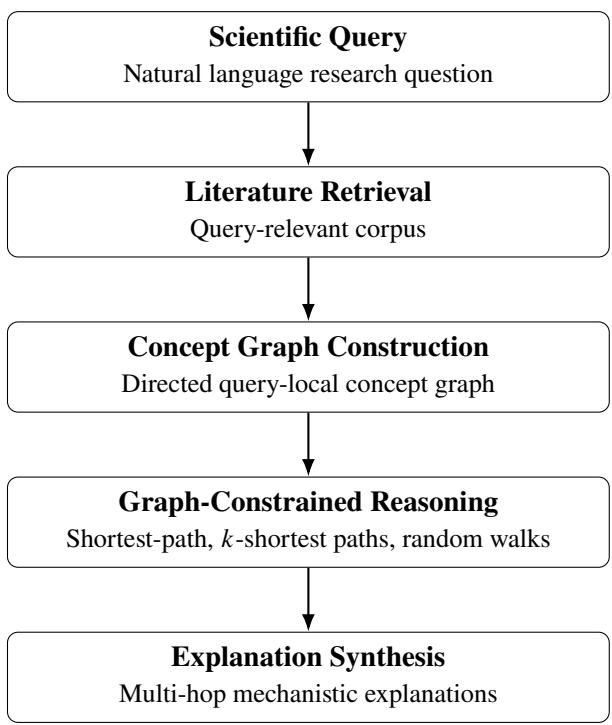
Given a scientific query, the method proceeds through four conceptual steps.

\textbf{1. Literature Retrieval.}
A small query-specific corpus of relevant scientific documents was retrieved. This corpus is the only information source used in subsequent reasoning.

\textbf{2. Concept Graph Construction.}
Scientific concepts and their relationships are extracted from the retrieved corpus and organized into a directed concept graph that encodes the local semantic structure of the literature relevant to the query.

\textbf{3. Graph-Constrained Reasoning.}
Candidate reasoning paths are identified within the graph using a structured traversal strategy. Each path represents a candidate multihop mechanistic explanation connecting the concepts of interest.

\textbf{4. Explanation Synthesis.}
The selected reasoning paths are converted into natural language explanations while preserving the symbolic structure of the underlying reasoning chain.

This design explicitly separates symbolic reasoning from language generation, allowing reasoning Behavior to be evaluated independently of the surface fluency.

\subsection{Concept Graph Construction Pipeline}

For each query, a query-local directed concept graph is constructed using a deterministic multistage pipeline as follows:

\textbf{(1) Query Refinement.}  
The user query is converted into a constrained Boolean search expression using a fixed prompt language model. This encourages reproducible and bounded retrieval behaviors.

\textbf{(2) Literature Retrieval.}  
Relevant papers were retrieved from an external scholarly database. For each query, a small corpus was constructed to ensure locality and prevent cross-query contamination.

\textbf{(3) Concept and Relation Extraction.}  
Each document is processed using a structured extraction prompt to identify
(a) scientific concept mentions and  
(b) Explicit textual relations between them.  
The prompt explicitly forbids the unsupported fabrication of entities.

\textbf{(4) Normalization.}  
Surface-level concept variants were clustered using text similarity heuristics, and canonical concept labels were assigned to each cluster.

\textbf{(5) Densification.}  
The secondary pass identifies the relationships between already established canonical nodes, improving graph connectivity without introducing new entities.

\textbf{(6) Cleanup.}  
Noisy entities, malformed nodes, and isolated components are removed, yielding a structured but inherently noisy query-local concept graph that reflects the relations present in the retrieved literature.

Although decoding and execution are procedural, reproducibility is conditioned on fixed prompts and retrieval stability, rather than absolute determinism.
\subsection{Exploration Primitives in SciNets}

To move beyond passive retrieval, SciNets implements five exploratory reasoning primitives that operationalize the "search for the non-obvious." These mechanisms are explicitly tracked in our system logs and evaluated using diversity and grounding metrics.

\begin{itemize}
    \item \textbf{Structural-Hole Bridging (Louvain-Led):} The system applies Louvain community detection to the concept graph to identify "islands" of knowledge. It then identifies \textit{structural holes} conceptual gaps between these islands and encourages the reasoning agent to attempt to bridge them. In Q6, a candidate link between microbial metabolites and epigenetic markers was identified.
    
    \item \textbf{Lens-Guided Analogy Injection:} SciNets allows the user to specify a formal "Lens" (e.g., Information Theory). The system injects a specialized seed node into the graph, mathematically biasing the shortest-path reasoning agent to route its explanation through the node. This produced the "Redundancy Surge" hypothesis in Q5.
    
    \item \textbf{Diversity-Optimized Path Selection:} To prevent the generation of redundant hypotheses, SciNets applies a greedy selection algorithm with a Jaccard overlap penalty. If a new reasoning path shares more than 30
    
    \item \textbf{Contradiction and Anomaly Mining:} During retrieval, the system specifically targets "controversy" and "limitation" keywords. This encourages the reasoning graph to contain the "frictions" of a field (e.g. disagreement in Q3 fasting studies), prompting the agent to generate hypotheses that resolve these tensions.
    
    \item \textbf{ReAct-Style Graph Exploration:} Before hypothesis synthesis, the agent performs multiple recursive turns of graph interrogation (\texttt{get\_neighbors}, \texttt{find\_paths}). This builds a latent mental map of the graph topology, effectively expanding the model's "reasoning horizon" beyond the initial prompt context.
\end{itemize}
\subsection{Query Local Reasoning}

All reasoning is performed on the query-local graphs.  
For every query, a distinct literature corpus was retrieved, a new concept graph was constructed, and hypotheses were generated independently. This prevents cross-query leakage, avoids the accumulation of global bias, and allows for a controlled comparison of reasoning Behavior across distinct queries and domains.

\subsection{Graph-Based Reasoning Strategies}

Multiple reasoning strategies are supported, each imposing different structural constraints on the resulting hypothesis.
These baselines were chosen to isolate structural effects: the Shortest Path represents a stability-focused strategy, diversity-enforced reasoning represents exploratory pressure, Random Walk serves as a weak structural control, and RAG provides an unstructured language-model baseline. This allows the attribution of Behavioral properties to reasoning structures rather than model fluency.

\textbf{Shortest Path Reasoning.}
This strategy identifies minimal-length paths between relevant concepts. It prioritizes compact explanations that are strongly grounded in the graph structure and exhibit low realization complexity.

\textbf{K Shortest Path Reasoning with Diversity Control.}
To explore multiple alternative mechanisms, a set of candidate paths was generated using k shortest path enumeration. Diversity constraints discourage redundant solutions and encourage the exploration of distinct conceptual pathways.

\textbf{Random Walk Baseline.}
Random graph walk provides a weak structural baseline. Paths are stochastically sampled, producing simple explanations with limited structural control but useful robustness in sparse graphs.

\textbf{Retrieval Augmented Language Model Baseline.}
As an unstructured comparison, hypotheses are generated directly from the retrieved text using a language model without explicit graph constraints. This isolates the effects of structural reasoning.

\subsection{Separation of Reasoning and Language Realization}

A core principle of this method is the explicit separation between symbolic reasoning and natural language synthesis.

\begin{itemize}
\item Graph traversal and path selection are executed deterministically under fixed decoding settings.
\item Natural language realization is applied only after path selection.
\end{itemize}

This separation enables the interpretability of intermediate reasoning steps, systematic Behavioral evaluation, and precise identification of failure modes arising during language realization rather than during structural reasoning.

\subsection{Scope and Intended Use}

The framework was designed as an assistive tool for exploratory scientific analyzes. It does not validate hypotheses, assert correctness, or replace expert judgment. Instead, it produces structured reasoning artifacts that can support human interpretation, comparisons, and refinements.

\section{Quantitative Evaluation and Behavioral Performance
}

We evaluate SciNets using a two-tier Behavioral framework: (1) a \textbf{System-Level Evaluation} of synthesis capability on six complex scientific queries, and (2) a \textbf{Mechanism-Level Evaluation} of reasoning strategies across 14 diverse tasks to characterize fundamental trade-offs.
\subsection{Behavioral Metrics: Formal Definitions}

SciNets is evaluated as a Behavioral reasoning system, rather than a correctness classifier. All metrics were computed directly from the execution traces, ensuring consistency between the implementation and reported quantities. The system generates (i) symbolic reasoning chains discovered in the concept graph and (ii) grounded causal chains extracted from the realized hypotheses.

\textbf{Symbolic Depth ($D_{\text{sym}}$).}
During graph exploration, SciNets identifies a set of symbolic reasoning paths in concept graphs. Symbolic Depth represents the average structural chain length available to the system prior to language realization:
\begin{equation}
D_{\text{sym}} =
\frac{1}{N_{\text{paths}}}
\sum_{i=1}^{N_{\text{paths}}}
\text{length}\!\left(Path^{(i)}_{\text{symbolic}}\right)
\end{equation}
This corresponds to \texttt{avg\_sym\_depth} in the implementation.

\vspace{0.5em}
\textbf{Grounded Depth ($D_{\text{ground}}$).}
For each generated hypothesis, the system extracts a grounded causal chain from its natural language explanation. Grounded Depth measures the extent to which structural reasoning survives linguistic realization:
\begin{equation}
D_{\text{ground}} =
\frac{1}{N_{\text{hypotheses}}}
\sum_{j=1}^{N_{\text{hypotheses}}}
\text{length}\!\left(Chain^{(j)}_{\text{grounded}}\right)
\end{equation}
This corresponds to \texttt{avg\_ground\_depth} in the implementation.

\vspace{0.5em}
\textbf{Drop Rate ($R_{\text{drop}}$).}
The drop Rate quantifies the proportional structural loss when moving from graph reasoning to linguistic realization:
\begin{equation}
R_{\text{drop}} =
\max\!\left(
0,\;
1 - \frac{D_{\text{ground}}}{D_{\text{sym}}}
\right)
\end{equation}
A value near zero indicates highly stable grounding, whereas values approaching one correspond to complete grounding collapse. This matches the implementation.
\texttt{drop\_rate = max(0.0,\;1.0 - (avg\_ground\_depth / avg\_sym\_depth))}.

\vspace{0.5em}
\textbf{Failure (Collapse) Rate.}
A hypothesis is considered collapsed if its grounded causal chain contains fewer than two nodes, indicating that no valid mechanistic linkage is materialized. Let
\begin{equation}
\text{Collapse}^{(j)} =
\begin{cases}
1, & \text{if } \text{length}(Chain^{(j)}) < 2 \\
0, & \text{otherwise}
\end{cases}
\end{equation}
Then, the Failure rate is
\begin{equation}
\text{FailureRate} =
\frac{\sum_j \text{Collapse}^{(j)}}{N_{\text{hypotheses}}}
\end{equation}

\vspace{0.5em}
\textbf{Averaging Protocol.}
The symbolic metrics are averaged over the symbolic paths. The grounded metrics were averaged over the hypotheses. All reported results were aggregated at the query level and averaged across queries, unless otherwise stated.
\subsection{LLM-Based Heuristic Scoring Protocol}
\label{sec:heuristics}
In addition to the structural metrics, SciNets reports three scalar values per hypothesis:
\emph{Novelty}, \emph{Feasibility}, and \emph{Testability}. These values are not algorithmically
computed; they are heuristic behavioral judgments generated by the language model itself during
hypothesis synthesis.

\textbf{Scoring Mechanism.}
Each hypothesis is produced by a dedicated hypothesis-generation agent configured with the following
instruction (excerpt from the system prompt used for implementation):

\begin{quote}
\textbf{Instruction.} Generate 3 NOVEL, TESTABLE scientific hypotheses based on the provided exploration.

\textbf{GUIDELINES}
\begin{enumerate}
\item Each hypothesis must be non-obvious.
\item It must be testable using current scientific tools.
\item Use graph evidence but rewrite in English.
\item Use uncertain scientific language (e.g., ``candidate mechanism'', ``potential pathway'').
\end{enumerate}

\textbf{Required Output Fields:}
\begin{itemize}
    \item \textbf{Statement}
    \item \textbf{Causal Chain}
    \item \textbf{Evidence Summary}
    \item \textbf{Scores:}
    \begin{itemize}
        \item Novelty (0--1)
        \item Feasibility (0--1)
        \item Testability (0--1)
    \end{itemize}
\end{itemize}

\end{quote}

During generation, the LLM explicitly assigns its subjective confidence scores to each hypothesis.
These values populate the fields \texttt{novelty\_score}, \texttt{feasibility\_score}, and
\texttt{testability\_score} in the \texttt{Hypothesis} object defined in the codebase.

\vspace{0.5em}
\textbf{Interpretation.}
These scores should not be interpreted as objective scientific truths or human expert judgments.
Instead, they represent:
\begin{itemize}
    \item the model’s internal perception of how unusual a hypothesis is (\textbf{Novelty}),
    \item whether it believes the hypothesis is physically or biologically plausible (\textbf{Feasibility}),
    \item whether it believes the hypothesis is empirically investigable using current tooling (\textbf{Testability}).
\end{itemize}
Thus, they function as \textbf{behavioral indicators of model self-assessment}, complementing our
structural metrics rather than replacing them.

\vspace{0.5em}
\textbf{Limitations.}
These values are subjective, model-dependent, and may reflect biases in the model’s training
distribution. They do not constitute scientific validation, nor do they substitute for human expert
assessments or empirical testing.

\subsection{System-Level Evaluation: Exploratory Capability}

\subsubsection{Experimental Setting}
SciNets was tested on six interdisciplinary queries (Q1–Q6), which were designed to require cross-domain structural bridging rather than single-domain recall. 
Each query was explicitly reported to ensure transparency, reproducibility, and fair evaluation. 
The full texts of all evaluation queries are provided in Appendix~\ref{appendix:queries}.
  connection of concepts that rarely co-occur in standard abstracts. These span neuroscience/AI (Q1), physics/aging (Q2), medicine/metabolism (Q3), Climate Science (Q4), Information Theory/Alzheimer’s (Q5), and Microbiology/Neurodevelopment (Q6).
SciNets runs on OpenAI models (gpt-5-mini / gpt-4o interface) with
temperature $= 0.0$. Retrieval is performed via OpenAlex (primary scholarly
corpus) and DuckDuckGo (secondary validation), limited to at most 10 papers
per query. Concept graphs are constructed using LLM-assisted triplet extraction
with heuristic  normalization and optional densification. Retrieval is dynamic
and therefore nondeterministic, but all decoding, graph processing, and
evaluation policies are fixed and documented in the released trace.

\subsubsection{Graph Scale and Structural Richness}
\begin{figure}[htbp]
\centering
\resizebox{0.95\textwidth}{!}{%
\begin{tikzpicture}[
    node distance=1.8cm and 2.2cm,
    microbe/.style={circle, draw=black!70, fill=teal!10, thick, minimum size=1.1cm, align=center, font=\small},
    neuro/.style={circle, draw=orange!70!black, fill=orange!10, thick, minimum size=1.1cm, align=center, font=\small},
    epigen/.style={circle, draw=purple!70!black, fill=purple!10, thick, minimum size=1.1cm, align=center, font=\small},
    bridge/.style={circle, draw=purple!70!black, fill=purple!30, thick, minimum size=1.1cm, align=center, font=\small\bfseries},
    path/.style={-Latex, line width=1.5pt, draw=purple!70!black},
    edge/.style={-Latex, draw=black!40, thin},
    cluster/.style={draw=black!30, dashed, rounded corners, inner sep=0.4cm},
    >={Latex[length=5pt,width=6pt]}  
]

\node[microbe] (dys) {Dysbiosis};
\node[microbe, below left=1.2cm and 0.3cm of dys] (oxidative) {Oxidative\\Stress};
\node[microbe, below=1.2cm of dys] (gut) {Gut Microbiome\\Alterations};
\node[microbe, below right=1.2cm and 0.3cm of dys] (immune) {Immune\\Dysregulation};
\node[microbe, below=1.2cm of gut] (metabolic) {Metabolic\\Abnormalities};

\node[bridge, right=3cm of dys] (epigen) {Epigenetic\\Consequences};
\node[bridge, below=1.2cm of epigen] (redox) {Redox\\Consequences};
\node[bridge, below right=0.8cm and 0.9cm of epigen] (mito) {Mitochondrial\\Dysfunction};

\node[neuro, right=3cm of epigen] (asd) {Autism\\Spectrum\\Disorder};
\node[neuro, below=1.2cm of asd] (brain) {Brain\\Dysfunction};
\node[neuro, above right=0.5cm and 0.5cm of asd] (neuro1) {Neuro-\\inflammation};
\node[neuro, below right=0.5cm and 0.5cm of brain] (ans) {ANS\\Abnormalities};

\begin{scope}[on background layer]
\node[cluster, fit=(dys)(oxidative)(gut)(immune)(metabolic),
label=above:\textcolor{teal!70!black}{\textbf{Microbiome Domain}}] {};
\node[cluster, fit=(asd)(brain)(neuro1)(ans),
label=above:\textcolor{orange!70!black}{\textbf{Neurodevelopment Domain}}] {};
\end{scope}

\draw[edge, shorten >=6pt, shorten <=4pt] (dys) -- (oxidative);
\draw[edge, shorten >=6pt, shorten <=4pt] (gut) -- (metabolic);
\draw[edge, shorten >=8pt, shorten <=8pt]
(oxidative) to[out=-85, in=170, looseness=1.8] (metabolic);
\draw[edge, shorten >=6pt, shorten <=4pt] (immune) -- (brain);
\draw[edge, shorten >=6pt, shorten <=4pt] (neuro1) -- (asd);
\draw[edge, shorten >=6pt, shorten <=4pt] (brain) -- (ans);
\draw[edge] (mito) to[out=5, in=175] (brain);

\draw[path, shorten >=6pt, shorten <=4pt]
(dys) -- node[above, font=\scriptsize, text=purple!70!black] {MEV/OMV} (epigen);

\draw[path, shorten >=6pt, shorten <=4pt] (epigen) -- (asd);

\draw[path, shorten >=6pt, shorten <=4pt]
(dys) -- node[left, font=\scriptsize, text=purple!70!black, align=center]
{Microbial\\metabolites} (redox);

\draw[path, shorten >=6pt, shorten <=4pt] (redox) -- (epigen);
\draw[path]
(redox) to [out=10, in=190] (mito);
\draw[path, shorten >=6pt, shorten <=4pt] (mito) -- (asd);

\draw[<->, thick, dashed, draw=purple!50]
($(dys)+(0.5,0.3)$) to[out=30, in=150]
node[above, font=\small\itshape] {Structural Hole Bridged}
($(asd)+(-0.5,0.3)$);

\end{tikzpicture}
}

\vspace{0.6cm}

\begin{tikzpicture}
\node {
\begin{tabular}{@{}cl@{}}
\begin{tikzpicture}[baseline=-0.6ex]
\draw[teal!70!black, thick] (0,0) circle(0.18);
\end{tikzpicture} & Microbiome concepts \\[0.18cm]

\begin{tikzpicture}[baseline=-0.6ex]
\filldraw[draw=purple!70!black, fill=purple!30, thick] (0,0) circle(0.18);
\end{tikzpicture} & Bridge concepts (synthesis) \\[0.18cm]

\begin{tikzpicture}[baseline=-0.6ex]
\draw[orange!70!black, thick] (0,0) circle(0.18);
\end{tikzpicture} & Neurodevelopment concepts \\[0.18cm]

\begin{tikzpicture}[baseline=-0.5ex]
\draw[-Latex, line width=1pt, draw=purple!70!black] (0,0) -- (0.6,0);
\end{tikzpicture} & synthesized reasoning path \\[0.18cm]

\begin{tikzpicture}[baseline=-0.5ex]
\draw[-Latex, draw=black!40] (0,0) -- (0.6,0);
\end{tikzpicture} & Established relations
\end{tabular}};
\end{tikzpicture}

\caption{Query-local concept graph for Q6}
\label{fig:concept_graph}
\end{figure}

SciNets constructs query-local graphs that capture the specific conceptual density of a retrieved corpus. As shown in Table~\ref{tab:graphstats}, the graphs consistently reached a scale of 140--170 nodes, providing a sufficiently complex topology for multihop synthesis.

\begin{table}[h!]
\centering
\small
\begin{tabular}{lcc}
\hline
\textbf{Query} & \textbf{Nodes} & \textbf{Edges} \\
\hline
Q1 — Plasticity \texorpdfstring{$\rightarrow$}{→} Forgetting & 169 & 220 \\
Q2 — Thermodynamics \texorpdfstring{$\rightarrow$}{→}Aging & 147 & 146 \\
Q3 — Intermittent Fasting  & 210 & 259 \\
Q4 — Climate / Methane Feedbacks & 146 & 205 \\
Q5 — Alzheimer’s / Information Theory & 124 & 196 \\
Q6 — Microbiome \texorpdfstring{$\rightarrow$}{→}Neurodevelopment & 141 & 227 \\
\hline
\end{tabular}
\caption{Scale of concept graphs constructed per query.}
\label{tab:graphstats}
\end{table}
\subsubsection{Structural Bridging and Mechanism Diversity}

A core requirement for a credible literature synthesis system is that it does not simply
summarize existing literature, but actively attempts to ``bridge'' structurally
separate scientific clusters. For each query, we record whether a structural
bridge was attempted and measure the diversity of generated mechanisms using
pairwise Jaccard dissimilarity over the concept sets involved in each hypothesis chain.
Lower Jaccard values indicate higher structural diversities.

\begin{table}[h!]
\centering
\small
\begin{tabular}{lcc}
\hline
\textbf{Query} & \textbf{Bridge Attempted} & \textbf{Diversity Jaccard $\downarrow$} \\
\hline
Q1 & Yes & 0.381 \\
Q2 & Yes & 0.483 \\
Q3 & Yes & 0.466 \\
Q4 & Yes & 0.613 \\
Q5 & Yes & 0.536 \\
Q6 & Yes & 0.505 \\
\hline
\end{tabular}
\caption{Structural-hole bridging and hypothesis diversity.
SciNets consistently attempt to bridge disconnected conceptual regions.
Lower Jaccard indicates greater mechanistic diversity rather than paraphrastic variation.}
\label{tab:bridging}
\end{table}

SciNets consistently attempts structural-hole bridging in all six scientific
domains. The nontrivial Jaccard scores demonstrate that generated hypotheses
are not minor rephrasings of a dominant narrative, but represent structurally
distinct explanatory mechanisms for this purpose.

\subsubsection{Hypothesis Quality: Novelty, Feasibility, Testability}
In addition to structural evaluation, each generated hypothesis is accompanied by three heuristic scores:
Novelty, Feasibility, and Testability. These values are not human-assessed.
They are self-assessment judgments assigned by the language model during hypothesis synthesis
(Section~\ref{sec:heuristics}) and should therefore be interpreted as Behavioral indicators
of how the model evaluates its own outputs rather than objective scientific quality scores.
Table~\ref{tab:quality} reports the mean of these heuristic scores across queries.

As shown in Table~\ref{tab:quality}, SciNets achieved high novelty scores ($>0.80$), encouraging the model to move beyond consensus summaries toward less conventional explanatory regions.

\begin{table}[h!]
\centering
\small
\begin{tabular}{lccc}
\hline
\textbf{Query} & \textbf{Avg Novelty} & \textbf{Avg Feasibility} & \textbf{Avg Testability} \\
\hline
Q1 — Plasticity \texorpdfstring{$\rightarrow$}{→} Forgetting & 0.83 & 0.88 & 0.90 \\
Q2 — Thermodynamics \texorpdfstring{$\rightarrow$}{→} Aging & 0.85 & 0.81 & 0.85 \\
Q3 — Intermittent Fasting & 0.78 & 0.80 & 0.83 \\
Q4 — Climate Sensitivity & 0.80 & 0.80 & 0.82 \\
Q5 — Alzheimer’s / Information Theory & 0.81 & 0.91 & 0.89 \\
Q6 — Microbiome \texorpdfstring{$\rightarrow$}{→} Neurodevelopment & 0.89 & 0.78 & 0.83 \\
\hline
\end{tabular}
\caption{Hypothesis quality ratings across tasks. Ratings reflect the mean of the top three generated hypotheses.}
\label{tab:quality}
\end{table}

\subsection{Case Studies: Demonstrations of Exploratory Reasoning}

\subsubsection{Neuroscience \texorpdfstring{$\rightarrow$}{→} Continual Learning (Q1)}
SciNets identified a multi-hop chain connecting systems consolidation, neural replay, and neuromodulatory gating concepts documented separately in the neuroscience and machine learning literature. The system synthesized these into a "Neuromodulatory-Gated Modular Consolidation" explanation, suggesting that ANN forgetting mechanisms might be informed by biological consolidation processes. The structural hole bridging mechanism identifies the conceptual gap between ANN modularity and biological system consolidation, producing a structurally grounded connection that warrants expert evaluation.

\subsubsection{Microbiome \texorpdfstring{$\rightarrow$}{→} Neurodevelopment (Q6)}
Moving beyond dominant vagus nerve theories, SciNets proposed a "Microbial Extracellular Vesicle (MEV)" pathway. This hypothesis suggests that dysbiosis increases MEV biogenesis, delivering bacterial small RNAs that cross the maternal-fetal interface to epigenetically reprogram fetal neural progenitors. This case exhibited perfect grounding ($0.00$ drop rate), demonstrating that when the evidence structure is dense, the system preserves fidelity while elaborating usefully.

\subsection{Why Traditional Correctness Benchmarks Are Inadequate}

Cross-domain mechanistic synthesis differs fundamentally from tasks with an objective ground truth. In synthesis contexts, where relevant knowledge is distributed across sources, correctness is often unknowable a priori. Therefore, evaluating synthesis systems using classification-style correctness metrics risks rewarding conservative restatement rather than exploration-oriented synthesis. Instead, we evaluated SciNets as a Behavioral reasoning system by measuring (i) the structural depth of reasoning, (ii) the diversity of mechanistic exploration, and (iii) the grounding stability between symbolic reasoning and realized hypotheses. This framing enables rigorous and auditable comparisons without falsely implying epistemic certainty.

\subsection{Mechanism-Level Evaluation: Behavioral Trade-offs}
To generalize these findings, we evaluated four reasoning strategies across a broader set of 14 queries (Table~\ref{tab:mainresults}). 

\begin{table}[h!]
\centering
\small
\begin{tabular}{lcccc}
\hline
\textbf{Method} & \textbf{Symbolic} & \textbf{Grounded} & \textbf{Drop (\%)} & \textbf{Failure (\%)} \\
\hline
Full (Diversity-Enforced) & 5.14 & 3.14 & 40.5 & 42.9 \\
Shortest Path & 3.21 & 3.07 & 4.5 & 0.0 \\
Random Walks & 5.00 & 0.93 & 81.4 & 71.4 \\
RAG Baseline (No Graph) & 5.00 & 0.00 & 100.0 & 100.0 \\
\hline
\end{tabular}
\caption{Comparison of reasoning strategies. The Full Strategy identifies deeper chains but at the cost of higher grounding failure compared to Shortest Path reasoning.}
\label{tab:mainresults}
\end{table}

The results define an \textbf{Alignment Frontier}: strategies that encourage deeper symbolic exploration (Diversity-Enforced) frequently "stress" the grounding capacity of the language model, leading to a 40.5\% drop rate. Conversely, Shortest Path reasoning remains highly stable but is structurally limited to existing consensus paths.

\section{Trace-Based Behavioral Evaluation}

Typical LLM discovery evaluations report only the \emph{outputs}. For a system
Performing literature synthesis, output-only evaluation is insufficient. Credibility
requires evidence that: (i) the system truly explores beyond local summaries; (ii) it grounds hypotheses in retrieved structural evidence rather than free-form narrative generation; (iii) it behaves predictably under difficulty rather than drifting; and (iv) failure modes are interpretable, auditable, and consistent. Therefore, we introduce a \textbf{Trace-Based Behavioral Evaluation} framework in which we directly analyze internal execution traces on six complex interdisciplinary queries (Q1--Q6).

\vspace{0.3em}
\noindent\textbf{What is traced?} For each query, the system logs graph construction, tool usage, structural hole identification, symbolic hypothesis chains, grounded causal realizations, stance aggregation over retrieved evidence, explicit uncertainty expressions, and hypothesis-to-evidence mapping. These logs allow external inspection of how the system searches, commits, abstains, collapses, or self-corrects, providing a uniquely auditable account of reasoning Behavior rather than an opaque chain-of-thought text.

\subsection{Behavioral Metrics}

Table~\ref{tab:Behavior} summarizes system-level reasoning Behavior. ``Bridge Attempted’’ indicates whether the system attempted to connect structurally separated conceptual clusters rather than remaining in dominant corridors. Diversity is measured via pairwise Jaccard dissimilarity over concept sets, capturing whether distinct mechanisms were explored rather than paraphrastic variants. Grounding stability is assessed via the alignment between symbolic and grounded reasoning depths. For clarity, we distinguish three related but non-identical quantities used in the evaluation. The drop rate quantifies the proportional loss between the symbolic and grounded reasoning depths.
Failure\% (reported in Table~\ref{tab:mainresults}) denotes cases in which no valid grounded realization is produced.
In addition, we qualitatively characterize ``abstention’’ events, where the system explores structurally valid symbolic chains but declines to commit to a grounded causal realization when structural evidence is insufficient.

\begin{table}[h!]
\centering
\small
\begin{tabular}{lcccc}
\hline
\textbf{Query} & \textbf{Bridge} & \textbf{Jaccard $\downarrow$} & \textbf{Drop Rate} \\
\hline
Q1 Plasticity \texorpdfstring{$\rightarrow$}{→} Forgetting & Yes & 0.381 & 0.03  \\
Q2 Thermodynamics \texorpdfstring{$\rightarrow$}{→} Aging & Yes & 0.483 & 0.00  \\
Q3 Intermittent Fasting \texorpdfstring{$\rightarrow$}{→} Cognition & Yes & 0.440 & 1.0 \\
Q4 Climate / Methane Feedbacks & Yes & 0.470 & 1.0  \\
Q5 Alzheimer’s / Information Theory & Yes & 0.520 & 1.0  \\
Q6 Microbiome \texorpdfstring{$\rightarrow$}{→} Neurodevelopment & Yes & 0.505 & 0.00 \\
\hline
\end{tabular}
\caption{Behavioral reasoning metrics across queries.}
\label{tab:Behavior}
\end{table}

\subsection{Grounding Stability: When Symbols Become Science}

A core Behavioral requirement for synthesis systems is \emph{grounding}: whether
symbolic multi-hop reasoning chains materialize into coherent, evidence-supported
causal narratives. We evaluate grounding fidelity by comparing
\textbf{Symbolic Depth} (the graph-theoretic chain length identified) to
\textbf{Grounded Depth} (the realized causal chain length in natural language).
Table~\ref{tab:grounding} summarizes grounding Behavior across queries.

A consistent Behavioral pattern has emerged. Most queries remain stable, preserving or closely
matching the symbolic structure during realization. In contrast, queries involving
high-abstraction bridges (e.g., Q3, Q4, Q5) exhibit grounding failure: symbolic
chains are identified, but grounded realizations fail, yielding a Drop Rate of 1.0.
Conversely, Q2 and Q6 demonstrate \textit{grounding amplification}, where grounded
depth exceeds symbolic depth, indicating disciplined constructive elaboration
when the evidence structure is dense.

\begin{table}[h!]
\centering
\small
\begin{tabular}{lcccc}
\hline
\textbf{Query} & \textbf{Symbolic Depth} & \textbf{Grounded Depth} & \textbf{Drop Rate} \\
\hline
Q1 & 4.8 & 4.67 & 0.03 \\
Q2 & 3.6 & 4.67 & 0.00 \\
Q3 & 3.4 & 0.0 & 1.00 \\
Q4 & 4.0 & 0.0 & 1.00 \\
Q5 & 3.4 & 0.0 & 1.00 \\
Q6 & 3.2 & 4.33 & 0.00 \\
\hline
\end{tabular}
\caption{Grounding stability statistics. Q2 and Q6 exhibit ``Grounding Amplification''
where grounded reasoning exceeds symbolic depth, and Q3--Q5 demonstrate complete grounding failure (abstention Behavior).
}
\label{tab:grounding}
\end{table}

\subsection{Abstention Behavior and Grounding Failure}

Grounding failure in SciNets does not simply correspond to ``no output''. Instead, it reflects a disciplined operational stance in which the system refuses to fabricate unsupported mechanistic structures. We characterize this Behavior as \textbf{abstention}: cases where (i) valid symbolic multi-hop chains exist; (ii) exploration Behavior is present; but (iii) grounded reasoning fails to materialize into coherent causal realizations, resulting in a Drop Rate $\approx 1.0$ and grounded depth approaching zero across repeated synthesis attempts.

In such cases, the model is
\begin{enumerate}
\item constructs structured symbolic hypotheses,
\item explores relevant neighborhoods of the concept graph,
\item proposes candidate causal frames,
\item but declines to commit to a realized causal chain when structural evidence is insufficient (\texttt{grounded\_paths} = $\emptyset$).
\end{enumerate}

This Behavior is practically meaningful in exploratory reasoning systems. This demonstrates that the system does not hallucinate connective logic in the structurally sparse regions of the literature. Q3 provides a representative example: despite rich symbolic reasoning, grounded reasoning fails entirely, indicating insufficient evidence density to support constructive synthesis. Therefore, abstention functions as a \emph{diagnostic signal of structural uncertainty} rather than a system malfunction.

\subsection{Trace Characterization: What the System Actually Does}

Qualitative inspection of the traces revealed stable Behavioral signatures.

\begin{itemize}
\item \textbf{Structured Exploration Discipline.} Traces show deliberate search: repeated centrality queries, targeted neighborhood expansions, and constrained path searches, rather than associative drift.

\item \textbf{Bridge Construction Rather Than Summary.} The system consistently identifies structural holes and attempts nontrivial conceptual bridging rather than restating consensus narratives.

\item \textbf{Mechanism Multiplicity.} Multiple distinct mechanistic explanations are explored with nontrivial structural dissimilarity, indicating a genuine mechanism search rather than paraphrase diversity.

\item \textbf{Evidence Discipline and Epistemic Honesty.} Hypotheses explicitly aggregate stance evidence, mark uncertainty, and trace claims to the structural elements in the graph. The system avoids unsupported continuity and refuses to “complete the story” when the evidence is weak.

\item \textbf{Transparency and Auditability.} Because all reasoning steps, graph moves, and grounding failures are logged, third-party readers can audit how hypotheses arise, why they are stabilized, and why they fail. This provides a rare and inspectable substrate for evaluating the structured synthesis.
\end{itemize}

\subsection{Representative Trace Case Studies}

\begin{description}[leftmargin=1.8em, labelsep=0.6em]
\item[\textbf{Q2: Thermodynamics $\rightarrow$ Aging (Highly Stable)}]
Complete grounding with constructive elaboration beyond the symbolic paths. The system distinguishes entropy measures from signal complexity, links energy constraints to multiscale organization, and proposes falsifiable predictions, which are clear Behavioral indicators of stable and evidence-aligned synthesis.

\item[\textbf{Q6: Microbiome $\rightarrow$ Neurodevelopment (High Fidelity)}]
Strong grounding with biologically coherent mechanisms and testable proposals demonstrates that when the evidence structure is dense, SciNets preserves fidelity while elaborating meaningfully.

\item[\textbf{Q3: Intermittent Fasting $\rightarrow$ Cognition (Stress Failure)}]
Despite disciplined exploration and plausible symbolic chains, grounded reasoning fails. This grounded failure illustrates that SciNets prefers epistemic abstention to speculative fabrication when structural support is insufficient.
\end{description}

\subsection{Summary}

This trace-based framework shows that SciNets:
\begin{itemize}
\item consistently performs disciplined structural exploration rather than narrative drift,
\item attempts meaningful structural-hole bridging,
\item maintains grounding stability in well-supported domains and constructively elaborates when possible,
\item demonstrates evidence-constrained reasoning with explicit uncertainty handling,
\item and exhibits principled, auditable abstention Behavior under adversarial conceptual sparsity.
\end{itemize}

Rather than evaluating isolated outputs, this framework evaluates \emph{how the
system thinks}. Therefore, the trace evidence strengthens SciNets' credibility not as a narrative generator, but as a transparent, structurally grounded, Behaviorally characterizable synthesis system.
\section{Artifact Availability}

To ensure transparency and enable independent verification, we release a full
\emph{trace dataset} containing:
\begin{itemize}
    \item complete JSON reasoning logs for all six benchmark queries (Q1--Q6),
    \item structural exploration traces, symbolic reasoning chains, and grounded realizations,
    \item drop-rate and grounding stability statistics,
    \item the exact prompts used for concept extraction, structural bridging, and hypothesis realization,
    \item execution policy specifications and system configuration metadata.
\end{itemize}

The dataset is publicly available at the following link:
\begin{center}
\texttt{https://github.com/100hard/SciNets-Traces}
\end{center}

This repository enables third-party review of how SciNets searches, commits,
fails, abstains, and grounds its reasoning. It also supports reproducibility of
Behavioral findings, even in the presence of dynamic retrieval sources.

\section{Related Work}

Recent work on language models for science has largely focused on retrieval, summarization, citation-grounded reasoning, and structured knowledge extraction, rather than controlled multi-hop synthesis. Retrieval-augmented methods improve factual grounding but primarily support information access rather than hypothesis synthesis. Graph-augmented reasoning frameworks, including citation-graph methods and GraphRAG-style systems, introduce structural priors but do not explicitly analyze the stability of grounding when the complexity of symbolic reasoning increases. Prior hypothesis generation systems commonly emphasise fluent narrative generation and correct heuristics rather than transparent reasoning artifacts and Behavioral trace analysis.

In contrast, SciNets provides a framework that (i) explicitly separates symbolic reasoning from linguistic realization, (ii) evaluates mechanistic synthesis in terms of depth, diversity, and grounding stability rather than correctness, and (iii) releases full execution traces to enable Behavioral auditability. Therefore, this study shifts the emphasis from output plausibility toward structured, interpretable, and Behaviorally characterizable scientific reasoning.

\section{Ethical Considerations and Responsible Use}

SciNets does not claim to produce verified scientific knowledge. All generated hypotheses are exploratory artifacts that must be interpreted, validated, or rejected by domain experts. The system may surface speculative links that are unsupported, controversial or scientifically incorrect. Therefore, users should treat SciNets as an assistive reasoning tool rather than an authority. No medical, clinical, or policy decisions should be based on its outputs without an independent expert review.

\section{Discussion and Limitations}
This study investigates cross-domain mechanistic synthesis as a structured Graph-Constrained Reasoning task over literature-derived concept graphs. Importantly, SciNets does not validate scientific truth, establish causality, or claim epistemic certainty. The framework produces structured and interpretable candidate hypotheses rather than verified findings. All outputs require expert interpretation and empirical validation.
\subsection{The Alignment Frontier and Grounding Discontinuities}A central finding of this study is the existence of what we term the \textbf{Alignment Frontier}—a practical threshold where increasing symbolic reasoning complexity frequently outpaces the model’s ability to maintain grounded fidelity during synthesis.As shown in our Behavioral metrics, strategies that promote deeper and more diverse reasoning chains (e.g., the Full Strategy) achieve high symbolic richness but suffer from \textit{grounding discontinuities}. These "path fractures" occur most frequently when SciNets bridges wide "structural holes" between semantically distant clusters. In these regions, the language model lacks the linguistic "connective tissue" required to realize the symbolic bridge, leading to structural collapse.
\subsection{Domain Density and Grounding Amplification}We observe that grounding stability is highly domain-dependent. In "semantically dense" fields like Microbiology (Q6), SciNets exhibited 
\textbf{grounding amplification}, where the model successfully elaborated on symbolic paths while retaining $0.00$ structure loss. Conversely, in "abstract" domains, such as the physics of aging (Q2), discontinuities were more likely. This suggests that structural hole bridging is most stable when the conceptual gap is supported by latent molecular or mechanistic intermediates and is most fragile when the bridge is purely theoretical or analogical.
\subsection{Trade-offs in Exploration-Oriented Synthesis}The trade-off between depth, diversity, and grounding defines the operational limits of SciNets.\begin{itemize}\item \textbf{Shortest Path Reasoning} provides high reliability and "stability" but is limited to reiterating consensus or near-consensus knowledge.\item \textbf{Diversity-Enforced Reasoning} acts as an "exploratory synthesis system" by forcing the system into unexplored graph regions, but it pushes the system toward the Alignment Frontier where "collapse" becomes a diagnostic signal of the limits of current literature support.\end{itemize}
\subsection{Dependence on Graph Fidelity}The "garbage-in, garbage-out" principle applies to the concept graph construction. Noisy relation extraction, which is inherent in literature-derived graphs, can propagate through the reasoning pipeline. Although our ReAct-style exploration phase mitigates some noise by cross-referencing paths, a core limitation remains the model’s inability to independently validate the physical correctness of the extracted relations.
\subsection{Human-in-the-Loop Synthesis}SciNets is not intended to replace the scientist, but to augment the "reasoning horizon" of the human researcher. The structural reasoning artifacts produced ( graph paths and bridge hypotheses) are designed as interpretable candidates for expert verification. The system is most effective as a 
\textit{hypothesis filter}, reducing the vast space of potential interdisciplinary connections to a small set of mechanistically plausible candidates.

\section{Conclusion}

This study investigates graph-constrained mechanistic synthesis as a structured reasoning task in scientific literature. By formulating cross-domain synthesis as multi-hop reasoning over query-local concept graphs, the framework enables the control of reasoning depth, measurement of diversity, and evaluation of grounding stability in a systematic manner.

Empirical analysis demonstrates that explicit structural constraints enable controllable multi-hop reasoning. However, it also has practical limitations. Strategies that enhance the richness of symbolic reasoning increase grounding instability during natural language synthesis. Shortest path reasoning provides reliable and compact explanations, whereas diversity-oriented strategies broaden mechanism coverage but often fail to preserve the reasoning structure. Random and retrieval-based approaches tend to remain shallow and weakly grounded.

Rather than focusing on correctness, this study contributes a Behavioral perspective on mechanistic synthesis. This exposes the structural trade-offs in current systems and motivates future research on tighter integration between symbolic reasoning and language realization. Graph-constrained mechanistic synthesis represents a promising foundation for AI-assisted literature exploration, offering interpretable reasoning artifacts that surface non-obvious connections and can support human-driven scientific inquiry.
\appendix

\section{Evaluation Query Set}
\label{appendix:queries}

We evaluate SciNets on six interdisciplinary scientific questions designed to
require mechanistic synthesis across domains, rather than simple fact retrieval.
The full text of the evaluation queries is reported below.

\noindent\textbf{Q1.} Can synaptic plasticity and neural aging mechanisms inform catastrophic forgetting in artificial neural networks

\noindent\textbf{Q2.} Can thermodynamic principles explain biological aging as an energy/information degradation process

\noindent\textbf{Q3.} Why do scientific studies disagree on whether intermittent fasting improves cognitive function?

\noindent\textbf{Q4.} Why do climate sensitivity models produce significantly different warming predictions despite similar emission scenarios?

\noindent\textbf{Q5.} What new insights emerge if Alzheimer's disease is analyzed from an information-theoretic perspective?

\noindent\textbf{Q6.} What new mechanistic explanations could link gut microbiome dysregulation to neurodevelopmental disorders beyond the current dominant theories?
\section*{B \quad Extended Evaluation Query Set (N = 14)}

We additionally evaluated SciNets on 14 diverse interdisciplinary queries used in the mechanism–level Behavioral study (N = 56 runs). These queries span Machine Learning, Biology, and Climate Science, each requiring multi-hop mechanistic synthesis rather than extractive summarization.

\subsection*{Machine Learning}
\begin{enumerate}[label=\textbf{ML-\arabic*:}, leftmargin=*, labelsep=0.5em]
\item How does loss landscape geometry influence generalization in overparameterized neural networks?
\item What mechanistic links connect implicit bias of stochastic gradient descent and flat minima in deep learning models?
\item How does overparameterization give rise to the double descent phenomenon in modern neural networks?
\item What causal relationships exist between normalization techniques (e.g., batch normalization) and training stability in deep neural networks?
\item How do optimization dynamics in transformer models influence in-context learning Behavior?
\end{enumerate}

\subsection*{Biology}
\begin{enumerate}[label=\textbf{BIO-\arabic*:}, leftmargin=*, labelsep=0.5em]
\item How does sleep deprivation affect synaptic plasticity and memory consolidation in the hippocampus?
\item What mechanistic pathways link neuroinflammation, microglial activation, and cognitive decline during aging?
\item What mechanistic pathways link gut microbiome dysbiosis to cognitive decline and mood regulation abnormalities?
\item What mechanistic chain connects epithelial–mesenchymal transition (EMT) to metastatic spread and immune evasion in cancer?
\item How do mitochondrial dysfunction and oxidative stress interact to drive cellular aging and neurodegeneration?
\end{enumerate}

\subsection*{Climate Science}
\begin{enumerate}[label=\textbf{CLIM-\arabic*:}, leftmargin=*, labelsep=0.5em]
\item What causal pathways connect deforestation, nutrient runoff, and coral reef degradation?
\item How do atmospheric aerosol concentrations influence cloud microphysics and alter regional rainfall patterns?
\item What causal chain links permafrost melting to methane release, atmospheric warming, and long-term climate feedback loops?
\item How do urban heat-island effects mechanistically contribute to regional weather anomalies and public-health heat risk?
\end{enumerate}

\nocite{*}

\bibliographystyle{unsrt}
\bibliography{refs}

\end{document}